\documentclass[letterpaper]{article} 
\usepackage{aaai2026}  
\usepackage{times}  
\usepackage{helvet}  
\usepackage{courier}  
\usepackage[hyphens]{url}  
\usepackage{graphicx} 
\usepackage{natbib}  
\usepackage{caption} 
\usepackage{algorithm}
\usepackage{algorithmic}

\usepackage{multirow}
\usepackage{graphicx}
\usepackage[dvipsnames,table]{xcolor}
\usepackage{float}

\definecolor{White}{rgb}{1.,0.,1.}
\definecolor{first}{rgb}{.8,.0,.0}
\definecolor{second}{rgb}{.0,.6,.0}
\definecolor{third}{rgb}{.0,.0,.8}

\definecolor{ceiling}{RGB}{214,  38, 40}
\definecolor{floor}{RGB}{43, 160, 4}
\definecolor{wall}{RGB}{158, 216, 229}
\definecolor{window}{RGB}{114, 158, 206}
\definecolor{chair}{RGB}{204, 204, 91}
\definecolor{bed}{RGB}{255, 186, 119}
\definecolor{sofa}{RGB}{147, 102, 188}
\definecolor{table}{RGB}{30, 119, 181}
\definecolor{tvs}{RGB}{160, 188, 33}
\definecolor{furniture}{RGB}{255, 127, 12}
\definecolor{objects}{RGB}{196, 175, 214}

\definecolor{car}{rgb}{0.39215686, 0.58823529, 0.96078431}
\definecolor{bicycle}{rgb}{0.39215686, 0.90196078, 0.96078431}
\definecolor{motorcycle}{rgb}{0.11764706, 0.23529412, 0.58823529}
\definecolor{truck}{rgb}{0.31372549, 0.11764706, 0.70588235}
\definecolor{othervehicle}{rgb}{0.39215686, 0.31372549, 0.98039216}
\definecolor{person}{rgb}{1.        , 0.11764706, 0.11764706}
\definecolor{bicyclist}{rgb}{1.        , 0.15686275, 0.78431373}
\definecolor{motorcyclist}{rgb}{0.58823529, 0.11764706, 0.35294118}
\definecolor{road}{rgb}{1.        , 0.        , 1.        }
\definecolor{parking}{rgb}{1.        , 0.58823529, 1.        }
\definecolor{sidewalk}{rgb}{0.29411765, 0.        , 0.29411765}
\definecolor{otherground}{rgb}{0.68627451, 0.        , 0.29411765}
\definecolor{building}{rgb}{1.        , 0.78431373, 0.        }
\definecolor{fence}{rgb}{1.        , 0.47058824, 0.19607843}
\definecolor{vegetation}{rgb}{0.        , 0.68627451, 0.        }
\definecolor{trunk}{rgb}{0.52941176, 0.23529412, 0.        }
\definecolor{terrain}{rgb}{0.58823529, 0.94117647, 0.31372549}
\definecolor{pole}{rgb}{1.        , 0.94117647, 0.58823529}
\definecolor{trafficsign}{rgb}{1.        , 0.        , 0.        }
\definecolor{otherstructure}{rgb}{0.98039215, 0.58823529, 0.}
\definecolor{otherobject}{rgb}{0.19607843, 1.        , 1.        }

\usepackage{amsmath}
\usepackage{amssymb}
\usepackage{booktabs}
\usepackage{adjustbox}
\usepackage[dvipsnames]{xcolor}
\usepackage{newfloat}
\usepackage{listings}
\DeclareCaptionStyle{ruled}{labelfont=normalfont,labelsep=colon,strut=off} 
\floatstyle{ruled}
\newfloat{listing}{tb}{lst}{}
\floatname{listing}{Listing}
\title{Towards 3D Object-Centric Feature Learning for Semantic Scene Completion}
\author{
    Weihua Wang\textsuperscript{\rm 1,2}\equalcontrib, 
    Yubo Cui\textsuperscript{\rm 1}\equalcontrib, 
    Xiangru Lin\textsuperscript{\rm 4}, 
    Zhiheng Li\textsuperscript{\rm 1}, 
    Zheng Fang\textsuperscript{\rm 1,2,3}\thanks{Corresponding author.}
}
\affiliations{
    \textsuperscript{\rm 1}Faculty of Robot Science and Engineering, Northeastern University, Shenyang, Liaoning, China\\
    \textsuperscript{\rm 2}National Frontiers Science Center for Industrial Intelligence and Systems Optimization, Shenyang, Liaoning, China \\
    \textsuperscript{\rm 3}The Key Laboratory of Data Analytics and Optimization for Smart Industry(Northeastern University), China \\
    \textsuperscript{\rm 4}The University of Hong Kong \\

    \{wangweihua,ybcui21, zhli24\}@stumail.neu.edu.cn,
    xiangru.lin@connect.hku.hk,\\
    fangzheng@mail.neu.edu.cn
}

\usepackage{bibentry}

\begin{document}

\maketitle

\begin{abstract}
Vision-based 3D Semantic Scene Completion (SSC) has received growing attention due to its potential in autonomous driving. While most existing approaches follow an ego-centric paradigm by aggregating and diffusing features over the entire scene, they often overlook fine-grained object-level details, leading to semantic and geometric ambiguities, especially in complex environments. To address this limitation, we propose Ocean, an object-centric prediction framework that decomposes the scene into individual object instances to enable more accurate semantic occupancy prediction. 
Specifically, we first employ a lightweight segmentation model, MobileSAM, to extract instance masks from the input image. Then, we introduce a 3D Semantic Group Attention module that leverages linear attention to aggregate object-centric features in 3D space. To handle segmentation errors and missing instances, we further design a Global Similarity-Guided Attention module that leverages segmentation features for global interaction. Finally, we propose an Instance-aware Local Diffusion module that improves instance features through a generative process and subsequently refines the scene representation in the BEV space.
Extensive experiments on the SemanticKITTI and SSCBench-KITTI360 benchmarks demonstrate that Ocean achieves state-of-the-art performance, with mIoU scores of 17.40 and 20.28, respectively.
\end{abstract}


\section{Introduction}
\label{sec:intro}
Semantic Scene Completion (SSC), also known as semantic occupancy prediction, has made significant progress in recent years. By partitioning the 3D space into voxels and predicting a semantic label for each voxel, SSC generates a dense and structured representation of the 3D environment. This fine-grained and enriched semantic detail can better serve downstream planning in autonomous driving. 
Compared to LiDAR-based approaches, vision-based methods are gaining increasing attention due to their lower cost and accessibility. By incorporating depth estimation, monocular vision-based SSC methods can infer 3D semantic information of the surrounding scene using only RGB images.


\begin{figure}[t]
    \centering
    \includegraphics[width=\linewidth]{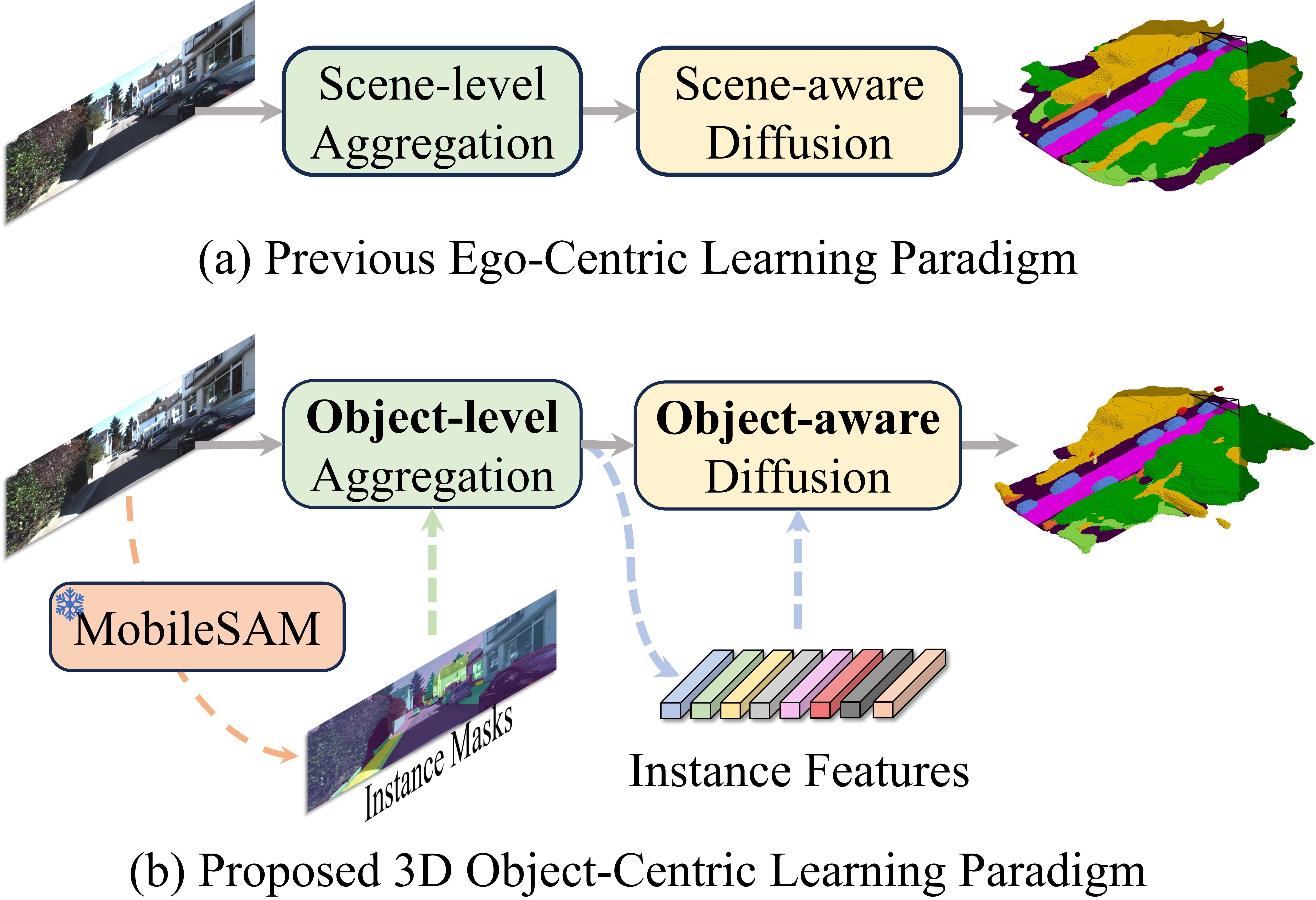}
    \caption{Comparison between our object-centric learning guided by MobileSAM and previous scene-level paradigms.
    }
    \label{fig:intro}
\end{figure}

To project 2D visual features into 3D space for prediction, most previous methods follow an ego-centric paradigm. This approach uses camera-ego relations to transform 2D features into 3D features and then diffuses these features across the entire scene.
For example, MonoScene~\cite{cao2022monoscene} lifts the multi-scale visual feature to 3D spaces with the camera parameters and utilizes 3D convolution to complete the 3D feature to make predictions. VoxFormer~\cite{li2023voxformer} selects a subset of voxels as \textit{query proposals} based on depth predictions, using cross-attention to aggregate features from the image and then employing self-attention to propagate this aggregated information to the entire 3D scene.
However, despite these achievements, this global paradigm does not adequately distinguish between different voxels, resulting in unintended interactions that usually cause semantic and geometric ambiguity.
Specifically, fusing features from different objects would lead to semantic confusion, while fusing features from empty and occupied areas also leads to geometric confusion.
As illustrated in Figure~\ref{fig:intro}(a), when multiple nearby cars are along the roadside, the empty spaces between the cars are mistakenly assigned features resembling those of the cars due to global fusion.
This makes it difficult to distinguish individual cars and leads to incorrect long-trailing predictions.

In contrast, object-centric learning paradigm has been explored to emphasize more detailed feature interactions, employing instance queries~\cite{jiang2024symphonize} or Gaussian points~\cite{huang2024gaussianformer} to implicitly represent objects and aggregate features from image features. Compared to the ego-centric learning paradigm, this paradigm demonstrates greater potential in aggregating object details.
However, these methods lack explicit object-level correspondences, hindering effective feature interaction at the object level and resulting in limited performance gains. 
Recently, large visual foundation models such as SAM~\cite{kirillov2023segment} have made significant strides in visual understanding, particularly in capturing fine-grained image details. Building on this, MobileSAM~\cite{zhang2023faster} adopts a lightweight design that enables the practical deployment of SAM in real-world downstream tasks.
This raises the question: \textit{Can these models be leveraged to provide finer details and enhance vision-based SSC predictions?}
However, leveraging these vision foundation models poses two major challenges: 1) the mask priors are constrained to the 2D image plane, which limits their capacity for comprehensive 3D feature interaction. 2) Mistakes and omissions in the prior masks may result in noticeable performance degradation.

To address these challenges, we introduce ocean, a 3D object-centric feature learning network for semantic scene completion, which integrates a SemGroup Dual Attention (SGDA) block and an Instance-aware Local Diffusion (ILD) module. 
The SGDA block incorporates two key components, 3D Semantic Group Attention (SGA3D) and Global Similarity-Guided Attention (GSGA), to enhance 3D semantic understanding.
Specifically, the SGA3D module facilitates object-level semantic aggregation by performing feature interactions within individual instances in 3D space, while GSGA leverages prior features to correct mask errors and reduce foreground omissions caused by imperfect masks.
Furthermore, the ILD module leverages instance-level features to reconstruct bird’s-eye view representations, which are then used to refine the entire scene feature for finer-grained semantic representation.

In summary, our contributions are as follows:
\begin{itemize}
    \item To address object-level semantic and geometric ambiguities in scenes, we propose Ocean, a novel approach that explicitly aggregates 3D object-level features and effectively enhances scene feature representations.
    \item To comprehensively utilize MobileSAM priors, we propose SGDA and ILD module, which are designed to aggregate object-centric features and diffuse instance-aware information throughout the scene, respectively.
    \item Experiments on the SemanticKITTI and SSC-Bench-KITTI360 show that Ocean achieves state-of-the-art performance with mIoU of 17.40 and 20.28, respectively.
\end{itemize}

\section{Related Works}
\label{sec:related_works}
\subsection{Vision-based 3D Perception}
The perception system plays a crucial role in autonomous driving. Compared to LiDAR-based methods, vision-based methods~\cite{philion2020lift, li2022bevformer} offer advantages in terms of lower cost and easier deployment, thus have garnered increasing attention.
LSS~\cite{philion2020lift} lifts 2D image features into 3D by predicting a per-pixel depth distribution and computing its outer product with the features.
However, LSS is sensitive to depth. In contrast, 
BEVformer~\cite{li2022bevformer} directly defines grid-shaped BEV queries to aggregate the BEV features. 
FB-BEV~\cite{li2023fb} analyzes the limitations of forward and backward projection methods and proposes a unified forward-backward paradigm to address them.
Nowadays, with the development of visual research, visual perception has made tremendous progress in detection~\cite{qi2024ocbev, zhang2025geobev}, segmentation~\cite{lu2025toward}.

\subsection{Semantic Scene Completion}
MonoScene~\cite{cao2022monoscene} is the first to leverage monocular images for the 3D SSC task. It lifts 2D visual features to 3D space and employs a 3D U-Net to extract voxel features for the final prediction. 
Moreover, TPVFormer~\cite{huang2023tri} introduces a tri-perspective view (TPV) and utilizes cross-view hybrid attention to enable interaction of features across the TPV planes. 
OccFormer~\cite{zhang2023occformer} decomposes 3D feature processing into local and global scales to enable long-range and dynamic interactions.
Voxformer~\cite{li2023voxformer} incorporates depth priors and adopts a sparse-to-dense approach for 3D feature interactions. 
Building upon these works,
MonoOcc~\cite{zheng2024monoocc} improves 3D occupancy via cross-attention and temporal distillation.
CGFormer~\cite{yu2024context} addresses query ambiguity with a context and geometry aware design.
LOMA~\cite{cui2025loma} uses a VL-aware generator and Tri-plane Fusion to boost language integration and global 3D modeling.
To reduce the heavy computation burden,
SparseOcc~\cite{liu2024fully} decouples semantics and geometry, introducing a sparse and efficient framework.
OctreeOcc~\cite{lu2024octreeocc} addresses dense-grid limitations by adaptively utilizing octree representations.
GaussianFormer~\cite{huang2024gaussianformer} first employs sparse gaussian representations for feature aggregation.
Different from the above methods, Symphonies~\cite{jiang2024symphonize} uses learnable instance queries to iteratively aggregate 3D features in an object-centric pipeline.
This paper adopts object-centric feature learning but uniquely leverages object priors for explicit object-level interactions.

\subsection{Object-Centric Learning}

Object-centric learning aims to represent the entire scene by utilizing multiple distinct objects within it. AIR~\cite{eslami2016attend} models each object using appearance, position, and presence, and these components are combined to form the final scene. 
In contrast, MONET~\cite{burgess2019monet} use spatial mixture models for scene representation, but are computationally inefficient. To solve this, Slot Attention~\cite{locatello2020object} develops a grouping strategy to create distinct slot representations.
However, slot counts can’t handle varying object numbers. AdaSlot~\cite{fan2024adaptive} dynamically adjusts slots based on instance complexity. Inspired by slot-based methods, we propose an instance-aware feature diffusion for 3D object-centric learning.
\begin{figure*}[t]
    \centering
    \includegraphics[width=\linewidth]{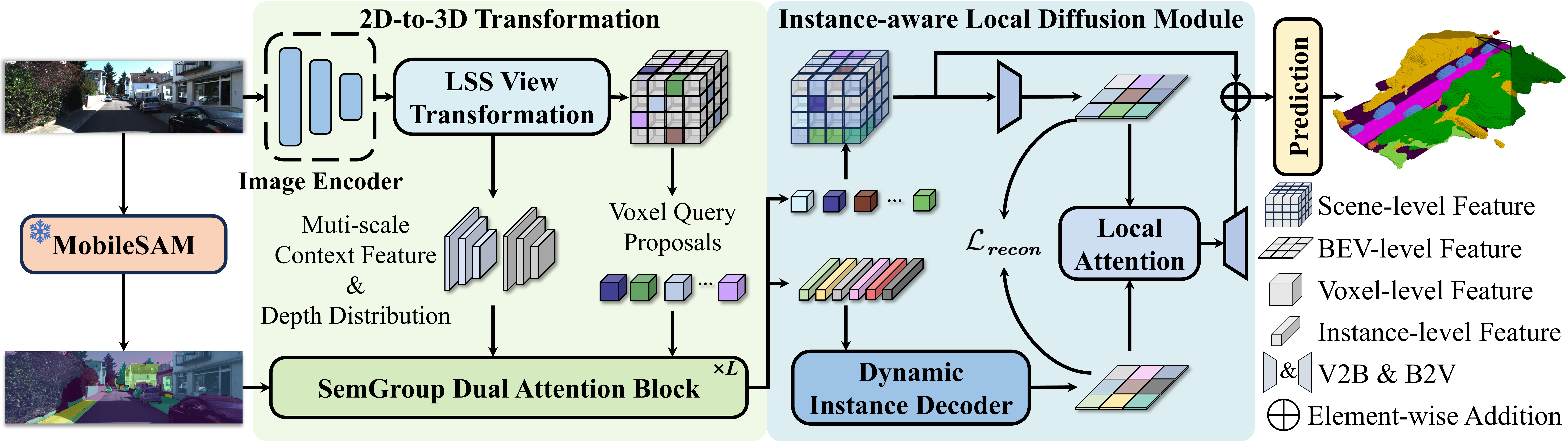}
    \caption{Overview of the proposed Ocean architecture. Given the monocular image as input, we first extract visual features using an image encoder and lift them into 3D space following LSS. To enable object-centric feature learning, we segment the scene using MobileSAM and design the SGDA block to aggregate features through both local and global attention. Furthermore, we propose the ILD module to refine the overall scene representation by incorporating instance-level features.}
    \label{fig:framework}
\end{figure*}
\section{Methods}
\subsection{Overview}
The overall architecture of the proposed method is shown in Figure~\ref{fig:framework}. 
Given a monocular image $\mathcal{I} \in \mathbb{R}^{3 \times H \times W}$, the goal of camera-based 3D occupancy prediction is to reconstruct a 3D occupancy scene $\mathcal{O} \in \mathbb{R}^{X \times Y \times Z \times (M + 1)}$, where $M$ denotes the number of semantic classes.
First of all, we employ an image encoder to extract multi-scale visual features $\{\mathcal{F}^{(s)} \in \mathbb{R}^{\frac{H}{s} \times \frac{W}{s} \times C_1^{(s)}}\}$ at three different downsampling ratios $s \in \{4, 8, 16\}$.  
Then, each scale-specific feature $\mathcal{F}^{(s)}$ is further processed by two separate modules to predict the corresponding depth distribution $\mathcal{D}^{(s)} \in \mathbb{R}^{\frac{H}{s} \times \frac{W}{s} \times D}$ and context feature $\mathcal{X}^{(s)} \in \mathbb{R}^{\frac{H}{s} \times \frac{W}{s} \times C_2}$, where $D$ is the number of discretized depth bins. 
We compute the outer product between the predicted depth distribution $\mathcal{D}$ and the context feature $\mathcal{X}$, and then apply voxel pooling~\cite{philion2020lift} to lift the 2D image features into a 3D voxel representation $\mathcal{V} \in \mathbb{R}^{x \times y \times z \times C_2}$, where $x, y, z$ denote the spatial dimensions of the 3D scene.  
This lifting operation is performed using the features at scale $s = 8$, which provides a good trade-off between spatial resolution and computational efficiency.
Following previous works~\cite{li2023voxformer, yu2024context}, we also use a pre-trained network~\cite{shamsafar2022mobilestereonet} to predict the depth map, and then select the 3D voxel queries $\mathcal{Q}\in \mathbb{R}^{N\times C_2} (0 \leq N \leq h \times w \times z)$, called query proposals, based on the depth prediction as follows:
\begin{equation}
    \mathcal{Q}=\mathcal{V}[M]
\end{equation}
where $M$ is a binary mask indicating whether the voxel is occupied based on the depth prediction.
Subsequently, to incorporate the object-level mask guidance from MobileSAM into the SSC task, we propose the SemGroup Dual Attention (SGDA) block, which aggregates query proposals with visual features at the object level. 
Furthermore, we introduce the Instance-aware Local Diffusion (ILD) module, which leverages instance priors to refine scene features through localized feature fusion.
Finally, the Semantic Scene Completion prediction is generated through a 3D prediction head.

\begin{figure}[t]
    \centering
    \includegraphics[width=0.84\linewidth]{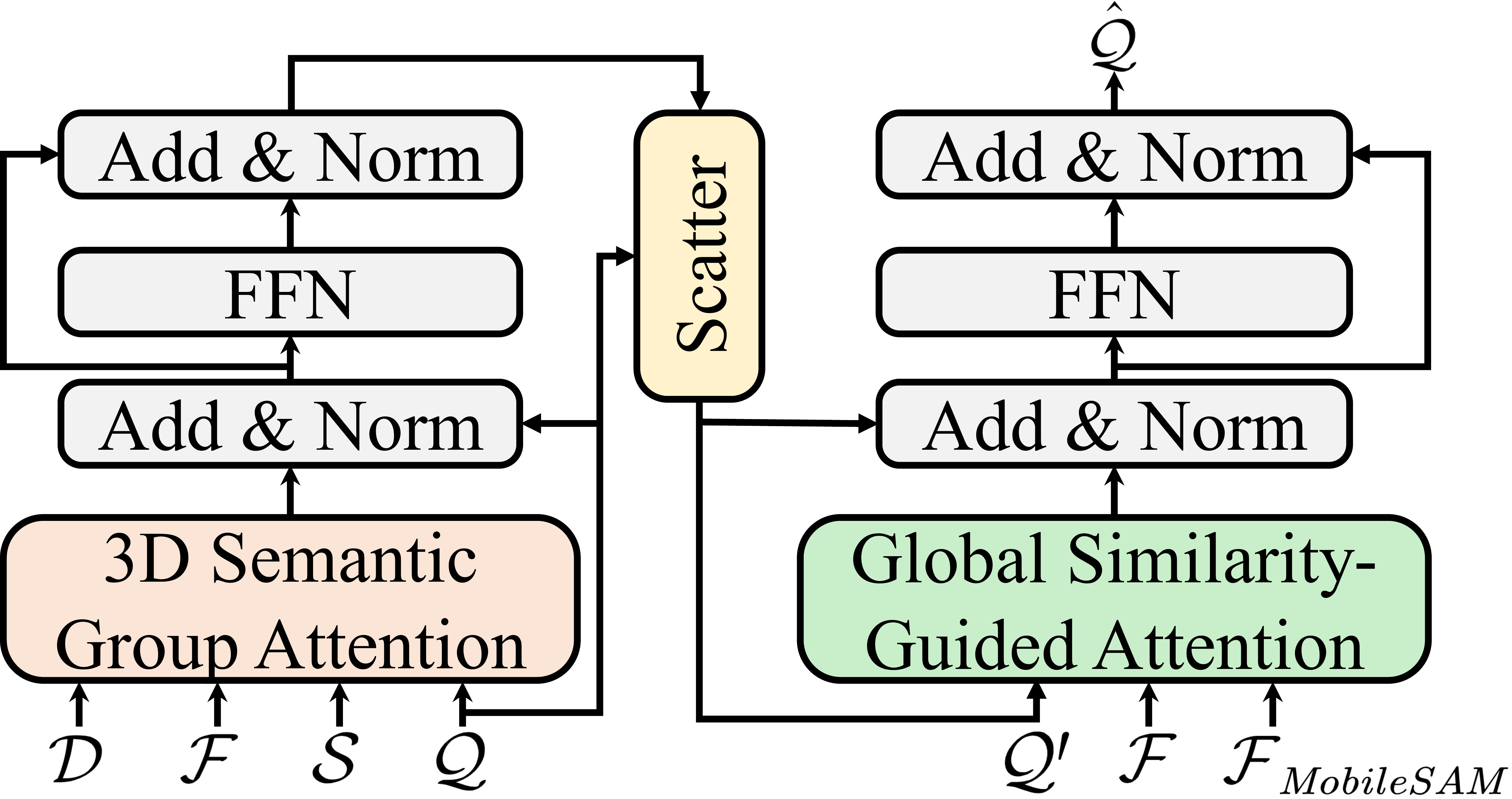}
    \caption{The details of SemGroup Dual Attention Block.
    }
    \label{fig:encoder}
\end{figure}

\subsection{SemGroup Dual Attention Block} 
To mitigate the ambiguity of object-level representations introduced by the ego-centric learning paradigm, we propose a novel object-centric framework. 
As shown in Figure~\ref{fig:encoder}, our approach introduces the SGDA block, which includes a 3D Semantic Group Attention module for object-centric local aggregation and a Global Similarity-Guided Attention module for context global interaction. 
This dual attention mechanism effectively captures both fine-grained local details and global contextual cues, significantly enhancing the quality and completeness of scene representations.

\subsubsection{Semantic Group Attention.}
\begin{figure}[t]
    \centering
    \includegraphics[width=0.9\linewidth]{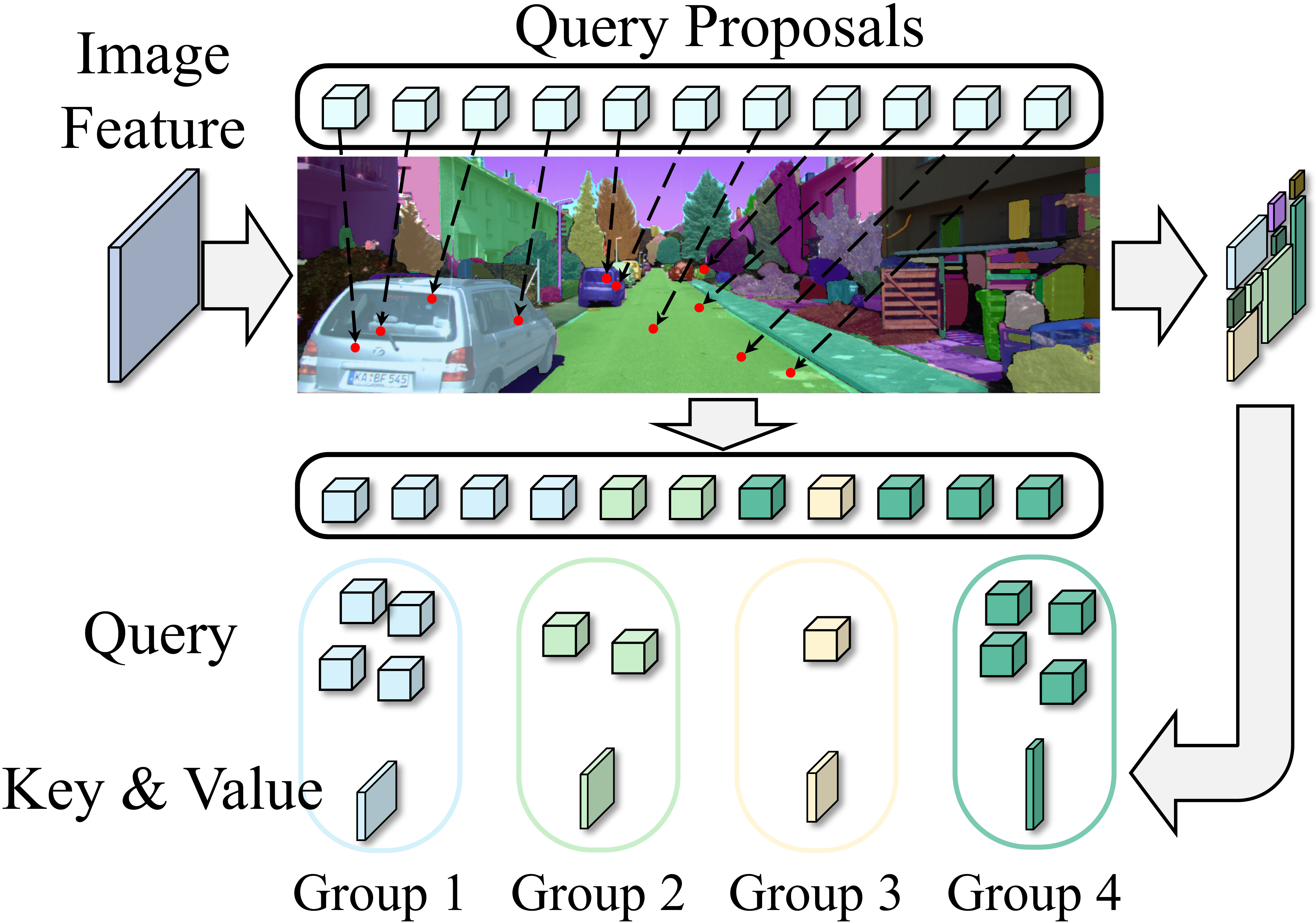}
    \caption{The Semantic Grouping. 3D query proposals are projected onto the image plane, assigned instance IDs via nearest-neighbor sampling, and clustered with image pixels of the same instance for aggregation using linear attention.
    }
    \label{fig:SG}
\end{figure}

Given a set of query proposals $\mathcal{Q} \in \mathbb{R}^{N \times C_2}$, our goal is to leverage visual features to enhance their representation for perception improvement. To this end, we first project the 3D query proposals onto the image plane using the camera's intrinsic and extrinsic matrices.
For a query located at $(x, y, z)$ in the ego-centric coordinate system, its corresponding 3D pixel coordinate $(u, v, d)$ in the image space is obtained as follows:
\begin{equation}~\label{equ:depth_compute}
d\left[\begin{array}{c}
u \\
v \\
1
\end{array}\right]=K\cdot \left[R|t\right]\cdot\left[\begin{array}{l}
x \\
y \\
z \\
1
\end{array}\right]
\end{equation}
where $K$ is the intrinsic matrix and $[R|t]$ are the camera extrinsic matrices, $d$ is the depth of the projected pixels.

After projection, we obtain 3D proposals that are aligned with the 2D image plane. These proposals are then used to aggregate corresponding instance-level image features on the image plane, facilitating an object-centric learning paradigm. However, acquiring accurate instance information from the image is inherently challenging, and learning features between image features of the same instance and their corresponding proposals remains a non-trivial task.
To this end, we adopt MobileSAM to extract per-pixel instance masks with high efficiency. As illustrated in Figure~\ref{fig:SG}, MobileSAM labels each pixel with an instance ID, generating a mask $\mathcal{S} \in \mathbb{R}^{H \times W \times 1}$ where pixels labeled as 0 denote background. Each 3D projection point is mapped to the image plane and assigned the instance ID of its nearest pixel.  
We use the nearest-neighbor approach to assign an instance ID to each projection point based on its projected coordinates $(u, v)$ to group the projection points with pixels. Pixels and projection points sharing the same instance are grouped into the same clusters. Projection points associated with background pixels are excluded from this process.

Meanwhile, higher-level features contain rich semantic information, while lower-level features preserve finer-grained texture details. 
For this purpose, we first downsample the instance masks to match the resolutions of the multi-scale features. 
We then group features across different scales based on the downsampled masks $\{\mathcal{S}_i\}^s_{i=0}$, and apply an attention mechanism within each group,  where the query proposals serve as the query $Q$ and the grouped pixel-level features serve as the key $K$ and value $V$, thereby aggregating multi-scale image features corresponding to the same instance ID. 
To further improve speed and memory efficiency, we employ scattered linear attention~\cite{he2024scatterformer} 
for feature aggregation, formulated as follows:
\begin{equation}
\tilde{Q}=\text{Concat}\left[
\frac{\varphi(Q^j) \sum_{i=1}^{m^j} \varphi(K_i^j)^{T} V_i^j}
     {\varphi(Q^j) \sum_{i=1}^{m^j} \varphi(K_i^j)^{T}}
\right]^M_{j=1}
\end{equation}
where $M$ denotes the number of instances, $m^j$ is the number of pixels in the cluster, and $\varphi(\cdot)$ denotes the kernel function.

\subsubsection{3D Extension with Depth Similarity.}
With the proposed SGA, we integrate the MobileSAM into SSC and perform object-centric learning.
However, this integration can only fuse features in 2D image space, limiting the model's ability to perceive features in 3D space, especially geometric perception. 
Meanwhile, due to the 2D limitation, the prior could not guide the 3D feature diffusion to capture detailed object information.
Thus, lifting the 2D prior to 3D space is essential for enhancing both semantic and geometric perception.

Inspired by DFA3D~\cite{DFA3D}, we extend SGA to 3D space by integrating depth information to enhance feature aggregation, particularly to capture richer geometric details. Specifically, within each cluster, the depth of grouped pixels is obtained from the predicted depth map $\mathcal{D}^{(s)}$, while the depth of grouped proposals is computed according to Equation~\ref{equ:depth_compute}.
However, pixel depth is represented as an uncertain probability distribution, whereas proposal depth is a deterministic computed value. This discrepancy hinders direct similarity computation between them.

Luckily, we observe that for each pixel, its predicted depth is represented by a softmax probability over multiple depth bins within a given depth range, which can also be interpreted as the pixel’s similarity to each depth bin. Meanwhile, we can also determine the corresponding depth bin for each projected depth of the proposal. 
Therefore, we can use the depth bin as a bridge to generate the depth similarity between the proposals and the pixels.
Specifically, for a cluster of $n$ pixels and $m$ proposals, we first gather the depth distribution of the pixels from the depth prediction. Next, for each proposal, we compute its projected depth value and gather the corresponding depth bin. We then extract the corresponding probability values from the pixel depth distribution, resulting in an $m \times n$ depth similarity matrix. 
This approach not only effectively represents the depth similarity between proposals and pixels but also fully leverages the predicted $\mathcal{D}$. Finally, we achieve the SGA3D as follows:
\begin{equation}
\tilde{Q}=\text{Concat}\left[
\frac{\varphi(Q^j) A^j \sum_{i=1}^{m^j} \varphi(K_i^j)^{T} V_i^j}
     {\varphi(Q^j) \sum_{i=1}^{m^j} \varphi(K_i^j)^{T}}
\right]^M_{j=1}
\end{equation}
where $A^j$ represents the depth similarity in the $j$-th cluster.

\subsubsection{Global Similarity-Guided Attention.}
Benefiting from the SGA3D module, our method enables instance-level feature aggregation in 3D space. However, SGA3D focuses on local feature interactions within individual instances, and the masks generated by MobileSAM may suffer from inaccuracies and missing instances. These limitations can significantly impact the overall prediction performance.

To address these limitations while maintaining the object-centric paradigm, we introduce the Global Similarity-Guided Attention (GSGA) module. It dynamically aggregates global image features to query proposals via a deformable attention mechanism, guided by global features from MobileSAM to reduce the impact of inaccurate masks.
Specifically, we compute the similarity between each query proposal and the MobileSAM intermediate features sampled at deformable offsets, and use these scores to filter and emphasize features belonging to the same object instance. This instance-aware guidance improves the relevance of feature aggregation, while the deformable mechanism enables the model to flexibly attend to informative regions beyond rigid segmentation boundaries. The final aggregated representation for each proposal is computed as:
\begin{equation}
\hat{Q} = \sum_k\left(G_k  W \mathcal{F}(p_q + \Delta p_k)\right) \odot A_k 
\end{equation}
Here, $\odot$ denotes element-wise multiplication, $k$ is the number of sampling points, $G_k$ denotes the similarity matrix between the query and the MobileSAM feature, $A_{k}$ represents the learnable attention weight at the $k$-th sampling location for a given query, and $\Delta p_{k}$ is the offset applied to the query position $p_q$. $W$ denotes the projection weight matrix.
\begin{figure}[t]
    \centering
    \includegraphics[width=\linewidth]{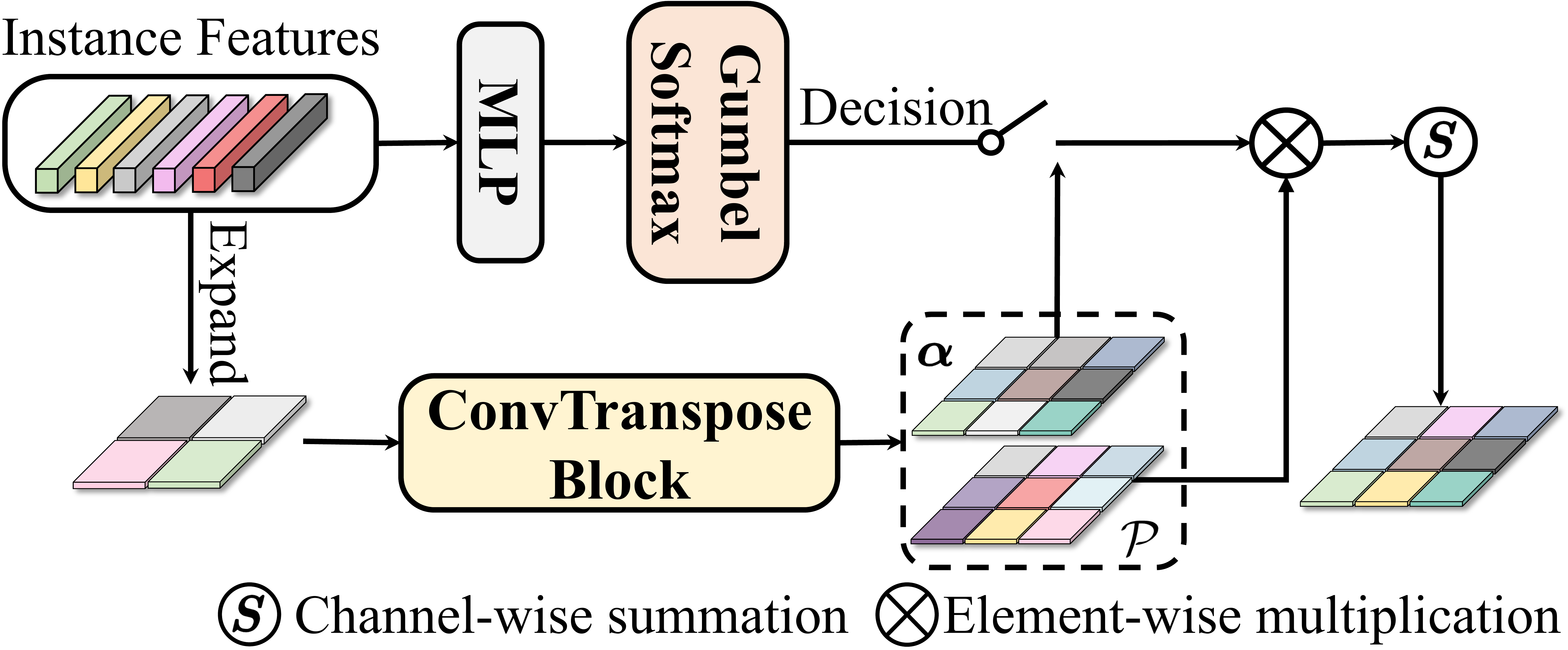}
    \caption{The details of the Dynamic Instance Decoder. Given the instance features, we reconstruct them into the scene-level BEV representation using a transposed convolutional decoder. Furthermore, we employ Gumbel Softmax to enable dynamic instance selection.}
    \label{fig:instance}
\end{figure}

\subsection{Instance-aware Local Diffusion Module}
Although the proposed SGDA block allows 3D query proposals to capture rich image features, many voxels still lack sufficient semantic information due to projection limitations. To address this, we introduce an Instance-aware Local Diffusion module that enhances spatial consistency by propagating features across the scene.

\subsubsection{Dynamic Instance Decoder.} 
To effectively leverage prior masks while preserving the instance-centric design, we sum grouped features across multiple scales based on instance masks to obtain aggregated instance-level representations. This simple summation incurs minimal computational cost and, due to its sensitivity to instance size, effectively preserves spatial characteristics at the instance level.

Given the extracted instance features, we further reconstruct corresponding BEV representations to enhance spatial and semantic understanding.
However, deriving fine-grained BEV features from independent instance representations is still a challenging task in complex and dynamic scenarios. To address this, we employ a generative strategy, using the voxel features from SGDA as guidance to dynamically generate BEV features enriched with instance information, as illustrated in Figure~\ref{fig:framework}. Specifically, we employ a lightweight deconvolution block to generate features with a dimension of $x\times y \times (C_2+1)$ from the instance features. As shown in Figure~\ref{fig:instance}, we also use a two-layer MLP along with Gumbel-Softmax to predict the one-hot decision $\mathcal{Z}$ for each instance feature.
Finally, we apply Softmax to normalize $\alpha$ based on $\mathcal{Z}$, then combine the weighted BEV features to reconstruct the scene BEV representation. The process could be represented as follows:
\begin{align}
    w_l=\frac{\exp(\alpha_l)}{\sum_{l=1}^L \exp(\alpha_l)}&, \hat{w}_l=\frac{\mathcal{Z}_l w_l}{\sum_{l=1}^L \mathcal{Z}_l w_l+\epsilon}\\ 
    \hat{\mathcal{P}} = \sum_{l=1}^{L}& \mathcal{P}_l\odot \hat{w}_l
\end{align}
where $L$ is the number of instances, $\epsilon$ is a small positive value for stable computation, $\mathcal{P}_i$ is $i$-th BEV feature.
Furthermore, inspired by the Slot-based method \cite{fan2024adaptive}, we introduce a reconstruction loss to constrain the generated BEV features, enhancing their quality and accelerating convergence. The loss function is defined as follows:
\begin{equation}
\mathcal{L}_{recon} = \sum_{i=1}^{N} \left\| \hat{\mathcal{P}}^{(i)} - \mathcal{F}_{bev}^{(i)} \right\|_2^2
\end{equation}
where $N$ represents the total number of elements , and $\mathcal{F}_{bev}^{(i)}$ denotes the feature corresponding to the Bird's Eye View (BEV) representation of the entire scene.

\subsubsection{Local Attention Refinement.}
Given the instance-aware BEV features, we use them to refine the aggregated BEV features.
Typically, neighboring voxels contain similar features. Therefore, we use window attention~\cite{liu2021swin} to refine the BEV features in a local manner. Specifically, we take the aggregated features as the query $Q$ and the instance-aware features as the key $K$ and value $V$. 
Finally, we use a 2D convolution layer to convert the refined BEV feature back to 3D shape and make predictions.
\begin{table*}[ht]
\centering
\setlength{\tabcolsep}{2.1pt}
{
\begin{tabular}{l|c c|ccccccccccccccccccccc}
    \toprule[.05cm]
    \textbf{Method}                               &
    \multicolumn{1}{c}{\textbf{IoU}}             &
    \textbf{mIoU}                                 &
    \rotatebox{90}{\textbf{road}}                 &
    \rotatebox{90}{\textbf{sidewalk}}                 &
    \rotatebox{90}{\textbf{parking}}                 &
    \rotatebox{90}{\textbf{other-grnd.}}                 &
    \rotatebox{90}{\textbf{building}}                 &
    \rotatebox{90}{\textbf{car}}                 &
    \rotatebox{90}{\textbf{truck}}                 &
    \rotatebox{90}{\textbf{bicycle}}                 &
    \rotatebox{90}{\textbf{motorcycle}}                 &
    \rotatebox{90}{\textbf{other-veh.}}                 &
    \rotatebox{90}{\textbf{vegetation}}                 &
    \rotatebox{90}{\textbf{trunk}}                 &
    \rotatebox{90}{\textbf{terrain}}                 &
    \rotatebox{90}{\textbf{person}}                 &
    \rotatebox{90}{\textbf{bicyclist}}                 &
    \rotatebox{90}{\textbf{motorcyclist}}                 &
    \rotatebox{90}{\textbf{fence}}                 &
    \rotatebox{90}{\textbf{pole}}                 &
    \rotatebox{90}{\textbf{traffic-sign}}                 &
    
    \\
    \midrule
    MonoScene   & 34.16          & 11.08          & 54.7          & 27.1          & 24.8          & 5.7           & 14.4          & 18.8          & 3.3          & 0.5          & 0.7          & 4.4          & 14.9          & 2.4           & 19.5          & 1.0          & 1.4          &0.4          & 11.1          & 3.3          & 2.1          \\
    TPVFormer     & 34.25          & 11.26          & 55.1          & 27.2          & 27.4          & 6.5          & 14.8          & 19.2          & 3.7 & 1.0          & 0.5          & 2.3          & 13.9          & 2.6           & 20.4          & 1.1          & 2.4          & 0.3          & 11.0          & 2.9          & 1.5          \\
    VoxFormer    & 42.95 & 12.20          & 53.9          & 25.3          & 21.1          & 5.6           & 19.8          & 20.8          & 3.5          & 1.0          & 0.7          & 3.7          & 22.4          & 7.5           & 21.3          & 1.4          & 2.6 & 0.2          & 11.1          & 5.1          & 4.9          \\
    OccFormer  & 34.53          & 12.32          & 55.9          & 30.3 & 31.5 & 6.5    & 15.7          & 21.6     & 1.2   & 1.5   &1.7    & 3.2          & 16.8    & 3.9    & 21.3    & 2.2   & 1.1   & 0.2          & 11.9    & 3.8   & 3.7  \\
    MonoOcc &- & 13.80 &55.2 &27.8 &25.1 & 9.7  &21.4 &23.2 & \underline{5.2}  & 2.2 &1.5 &5.4 &24.0 &8.7 &23.0 &1.7 &2.0 &0.2 &13.4 &5.8 & 6.4  \\
    Symphonies    & 42.19          &15.04 & 58.4 & 29.3  & 26.9  & 11.7 &24.7 & 23.6 & 3.2          & 3.6 & \textbf{2.6} & \underline{5.6} & 24.2 & 10.0 &23.1 & \textbf{3.2} & 1.9          & \textbf{2.0} &16.1&7.7 & 8.0 \\ 
    LOMA & 43.01       & 15.10 & 58.0 & 31.8   & 32.2   & 9.5 &25.3 & 24.9 & 4.1  & 1.7 & 1.7 & \textbf{6.4} & \underline{25.6} &8.7 & 24.7 & 1.4&1.7 & 0.6 & 16.8 & 6.5  &6.1 \\
    CGFormer  & \underline{44.41}          & 16.63         & 64.3         & 34.2 & \textbf{34.1} & 12.1   & \underline{25.8}       & 26.1     & 4.3   & \underline{3.7}  & 1.3  & 2.7         & 24.5    & \underline{11.2}    & 29.3 & 1.7   & \textbf{3.6}   & 0.4        & 18.7   & \underline{8.7}   & \underline{9.3}  \\
    HTCL & 44.23         & \underline{17.09}        & \underline{64.4}          & \underline{34.8}  & \underline{33.8} & \underline{12.4}  & \textbf{25.9}         & \textbf{27.3}    & \textbf{5.7}   & 1.8  & \underline{2.2}  &5.4         & 25.3    & 10.8    & \textbf{31.2}  &1.1  & \underline{3.1} & 0.9         & \textbf{21.1}  & \textbf{9.0}   &  8.3  \\
    Ocean (Ours)  & \textbf{45.62}  & \textbf{17.40} & \textbf{65.1} & \textbf{34.9}   & 33.7   & \textbf{12.8} & 25.4 &\underline{26.8} & \underline{5.2}  & \textbf{4.9}& 1.9 & \underline{5.6}  & \textbf{26.9} &\textbf{11.6} & \underline{30.7} & \underline{2.3} & 2.2 & \underline{1.7} &\underline{20.8} &\underline{8.7} &\textbf{9.5} \\
    \bottomrule[.05cm]
\end{tabular}
}
\caption{Quantitative results on SemanticKITTI test set. The best and second results are in \textbf{bold} and \underline{underlined}, respectively.}
\label{tab:sem_kitti_test}
\end{table*}

\begin{table*}[ht]
\centering
\setlength{\tabcolsep}{2.1pt}
{
\begin{tabular}{l|c c|ccccccccccccccccccccc}
        \toprule[.05cm]
        \textbf{Method}                               &
        \multicolumn{1}{c}{\textbf{IoU}}             &
        \textbf{mIoU}                                               &
        \rotatebox{90}{\textbf{car}}                 &
        \rotatebox{90}{\textbf{bicycle}}                 &
        \rotatebox{90}{\textbf{motorcycle}}                 &
        \rotatebox{90}{\textbf{truck}}                 &
        \rotatebox{90}{\textbf{other-veh.}}                 &
        \rotatebox{90}{\textbf{person}}                 &
        \rotatebox{90}{\textbf{road}}                 &
        \rotatebox{90}{\textbf{parking}}                 &
        \rotatebox{90}{\textbf{sidewalk}}                 &
        \rotatebox{90}{\textbf{other-grnd.}}                 &
        \rotatebox{90}{\textbf{building}}                 &
        \rotatebox{90}{\textbf{fence}}                 &
        \rotatebox{90}{\textbf{vegetation}}                 &
        \rotatebox{90}{\textbf{terrain}}                 &
        \rotatebox{90}{\textbf{pole}}                 &
        \rotatebox{90}{\textbf{traffic-sign}}                 &
        \rotatebox{90}{\textbf{other-struct.}}                 &
        \rotatebox{90}{\textbf{other-obj.}}                 &

        \\ 
        \midrule
        MonoScene   & 37.87          & 12.31        & 19.3        & 0.4        & 0.6        & 8.0         & 2.0         & 0.9        & 48.4        & 11.4        & 28.1        & 3.3        & 32.9        & 3.5        & 26.2        & 16.8        & 6.9         & 5.7        & 4.2         & 3.1         \\
        TPVFormer          & 40.22       & 13.64        & 21.6        & 1.1        & 1.4        & 8.2         & 2.6         & 2.4        & 53.0       & 12.0        & 31.1        & 3.8        & 34.8        & 4.8        & 30.1        & 17.5        & 7.5         & 5.9        & 5.5         & 2.7         \\
        VoxFormer           & 38.76         & 11.91        & 17.8        & 1.2        & 0.9        & 4.6         & 2.1         & 1.6        & 47.0        & 9.7         & 27.2        & 2.9        & 31.2        & 5.0        & 29.0        & 14.7        & 6.5         & 6.9        & 3.8         & 2.4         \\
        OccFormer           & 40.27      & 13.81        & 22.6        & 0.7        & 0.3        & 9.9         & 3.8         & 2.8        & 54.3        & 13.4        & 31.5        & 3.6        &  36.4 & 4.8        & 31.0        &  19.5 & 7.8         & 8.5        & 7.0         & 4.6         \\
        Symphonies        & 44.12    & 18.58 & \textbf{30.0} & 1.9  & \textbf{5.9} & \textbf{25.1} & \textbf{12.1} & \textbf{8.2} &  54.9  & 13.8 &  32.8 & \textbf{6.9} & 35.1        &  \underline{8.6}  &  38.3  & 11.5        & 14.0 & 9.6  & \textbf{14.4} & \textbf{11.3} \\  
     
        CGFormer     & \underline{48.07}    & \underline{20.05} & \underline{29.9} & \underline{3.4} &4.0  & \underline{17.6} & 6.8 & 6.7 &\textbf{63.9} &\textbf{17.2}& \underline{40.7} &\underline{5.5} & 42.7 &8.2 &38.8 &\textbf{24.9} &\underline{16.2} &\underline{17.5} &10.2 &6.8 \\
        SGFormer        & 46.35    & 18.30 & 27.8  & 0.9 & 2.6 & 10.7 & 5.7 & 4.3 &  61.0  & 13.2 &  37.0 & 5.1 & \underline{43.1}   &  7.5  &  \underline{39.0} & \textbf{24.9}       & 15.8 & 16.9  & 8.9 & 5.3 \\  
        
        Ocean (Ours)   & \textbf{48.19}   & \textbf{20.28} & 29.3 & \textbf{3.7} & \underline{4.6} & 15.1 & \underline{7.7} & \underline{6.8} & \underline{63.7} &  \underline{17.0}  & \textbf{40.9} & 5.0 &\textbf{43.7} & \textbf{8.9} & \textbf{39.2}    &  \underline{24.7} & \textbf{16.7} & \textbf{19.2}     & \underline{10.8} & \underline{8.3} &  \\
        \bottomrule[.05cm]
        \end{tabular}
}
\caption{Quantitative results on SSCBench-KITTI360 test set.  The best and second results are in \textbf{bold} and \underline{underlined}.}

\label{tab:kitti_360_test}
\end{table*}

\section{Experiments}

\subsection{Implementation Details}
Following previous works~\cite{cao2022monoscene}, we use EfficientNetB7~\cite{tan2019efficientnet} as our 2D backbone.
The SemGroup Dual Attention block consists of $3$ layers.
The query proposals from SGDA are scattered back to a 3D feature of size $128 \times 128 \times 16$ with $128$ channels, which is then processed by the ILD. 
We train Ocean on $4$ GeForce 3090 GPUs, using the AdamW optimizer~\cite{loshchilov2017decoupled} with an initial learning rate of 3e-4 and weight decay of 0.01.

\begin{figure*}[t]
    \centering
    \includegraphics[width=0.9\linewidth]{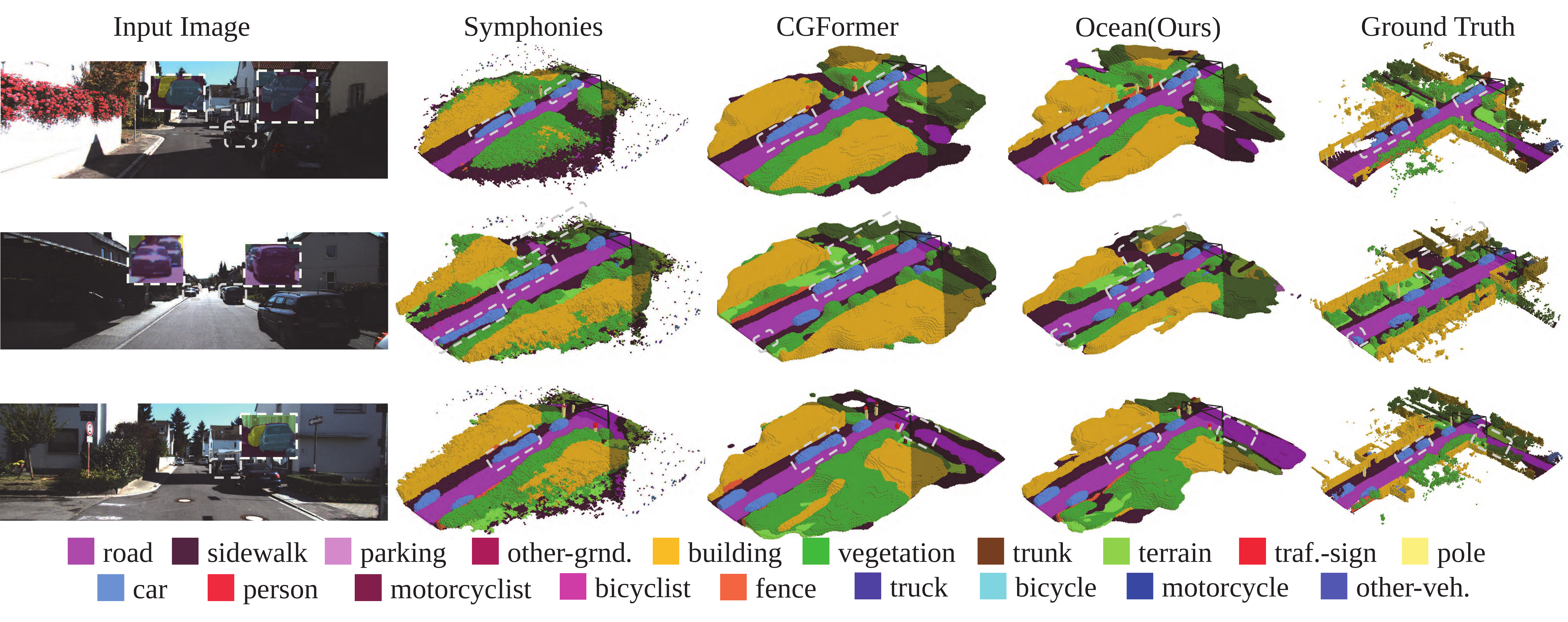}
    \caption{Qualitative visualizations on SemanticKITTI validation set. 
    }
    \label{fig:viz}
\end{figure*}

\subsection{Comparisons with the State-of-the-Art Methods} \label{Expriments:main results}
As shown in Table~\ref{tab:sem_kitti_test} and Table~\ref{tab:kitti_360_test}, our method achieves the best performance on both the SemanticKITTI~\cite{behley2019semantickitti} and the SSCBench-KITTI360~\cite{li2023sscbench} datasets.  
On the SemanticKITTI dataset, Ocean reaches an IoU of 46.40 and a mIoU of 17.39, outperforming CGFormer by 1.21 and 0.77, respectively. It also surpasses HTCL~\cite{li2024hierarchical}, which utilizes temporal information, demonstrating the effectiveness of our instance-level feature aggregation. 
Additionally, the SemanticKITTI validation visualizations in Figure~\ref{fig:viz} clearly demonstrate that our method achieves superior capability in capturing fine-grained scene structure, especially for objects with strong edge or contour cues, such as cars and roads.
On the SSCBench-KITTI360 dataset, Ocean exceeds CGFormer by 0.12 in IoU and 0.23 in mIoU, and further outperforms SGFormer~\cite{guo2025sgformer}, which leverages additional remote sensing modalities. These results confirm the advantages of our object-centric 3D scene understanding approach.
On both datasets, Ocean achieves outstanding performance in recognizing objects with consistent surface information, such as the road, sidewalks, and even smaller items such as poles and traffic signals, thanks to the guidance of prior knowledge and the design of the SGA3D algorithm. For objects with more complex surface information, like bicycles and vegetation, which are difficult for segmentation algorithms to distinguish, Ocean also delivers superior results, as the GSGA algorithm reduces reliance on prior information, mitigating information loss and segmentation errors.

\begin{table}[t]
\centering
\setlength{\tabcolsep}{1mm}
{
\begin{tabular}{c|cc|cc|cc}
\toprule[.05cm]
\multirow{2}{*}{}  & \multicolumn{2}{c|}{\textbf{SGDA}} & \multicolumn{2}{c|}{\textbf{ILD}} & \multicolumn{2}{c}{\textbf{Performance}} \\ 
\cline{2-3} \cline{4-5} \cline{6-7}
& \textbf{SGA3D}  & \textbf{GSGA}      & \textbf{LA}         & \textbf{DID}        & \textbf{IoU ↑}         & \textbf{mIoU ↑} \\
\midrule
   
M0 &     &     &        &      & 44.62  &   15.80       \\
M1 &   \checkmark  &     &   &       &   45.77    &  16.49         \\
M2    &  \checkmark  & \checkmark  & &  &  46.01  &   16.80   \\  

M3  &   \checkmark           &  \checkmark   &  \checkmark&   &  45.39 &  17.10   \\   
   
M4  &  \checkmark & \checkmark & \checkmark & \checkmark        &   \textbf{46.40} & \textbf{17.39} \\ 
\bottomrule[.05cm]
\end{tabular}}
\caption{Ablation study on each module in Ocean.}
\label{tab:main_module}
\end{table}

\subsection{Ablation study} \label{Expriments:ablation study}
To analyze the effectiveness of each module, we conduct ablation studies on the SemanticKITTI validation set.

\noindent\textbf{Main Modules.} Table~\ref{tab:main_module} presents an ablation study comparing the core modules of our method. 
Compared to our baseline, the introduction of prior knowledge and the application of SGA3D result in an improvement of 1.15/0.69 in IoU/mIoU for M1, demonstrating the effectiveness of our design. However, due to the mistakes and omissions in the prior masks, the model has not fully learned the scene information. Therefore, M2 further incorporates GSGA and achieves 46.01/16.80 in IoU/mIoU. This result demonstrates that aggregating features at the instance level feature in 3D space can effectively enhance both semantic and geometric perception.
Furthermore, after incorporating local feature diffusion, mIoU improved by 0.3. However, due to the voxel2bev operation, which enhances the instance's semantic information, some geometric information is inevitably lost. 
To address this issue, we further incorporate the Dynamic Instance Decoder (DID), which adopts a generative paradigm to enhance the overall scene semantic information and partially mitigate the loss of geometric details.

\begin{table}[t]
\centering
\setlength{\tabcolsep}{1mm}
{
\begin{tabular}{c|cccc}
\toprule[.05cm]
\textbf{Method} & \textbf{IoU ↑} & \textbf{mIoU ↑} & \textbf{Param} & \textbf{FLOPs}  \\
\midrule
SGA3D(Ours)   & 46.40   & \textbf{17.39} &  0.280M  & 1.333G \\
SGA3D w/o Ms.  & \textbf{46.44} & 16.64  & 0.280M  & 1.298G  \\
SGA3D w/o Ext.  & 45.84 & 17.30  & 0.280M  & 1.333G  \\
DFA2D  &  46.14 & 16.39  & 0.202M & 2.231G  \\
DFA3D  &  46.15 & 16.97  & 0.330M & 2.807G  \\
\bottomrule[.05cm]
\end{tabular}
}
\caption{Ablation study on 3D Semantic Group Attention.}
\label{tab:sga}
\end{table}




\begin{table}[t]
\centering
\setlength{\tabcolsep}{1mm}
\begin{tabular}{c|cc}
\toprule[.05cm]
\textbf{Method}  & \textbf{IoU ↑} & \textbf{mIoU ↑} \\
\midrule

Full Model(Ours)    & \textbf{46.40} & \textbf{17.39}\\ 
w/o GSGA        & 46.13 & 16.98 \\
GSGA w/o Sim. & 46.07 & 16.36 \\

\bottomrule[.05cm]
\end{tabular}
\caption{Ablation study in Global Similarity-Guided Attention.}
\label{tab:SGDFA}
\end{table}

\begin{table}[t]
\centering
{
\begin{tabular}{c|cc}
\toprule[.05cm]
\textbf{Method} & \textbf{IoU ↑} & \textbf{mIoU ↑} \\
\midrule
 Direct Sum  & 45.79 &   16.45  \\
  Softmax-Weighted    & 45.89 & 16.88  \\
 Sigmoid-Weighted    & 46.30 & 16.92    \\
 Dynamic Selection (Ours)   &\textbf{46.40}     & \textbf{17.39}    \\ 
\bottomrule[.05cm]
\end{tabular}
}
\caption{Ablation study in Dynamic Instance Decoder.}
\label{tab:instance_fusion}
\end{table}

\noindent\textbf{3D Semantic Group Attention.} 
As shown in Table~\ref{tab:sga}, we first evaluate the two core components of SGA3D. Without(w/o) the multi-scale image features, the model experiences a significant drop in semantic understanding, while removing the 3D extension design leads to a noticeable decline in geometric reasoning, demonstrating the distinct and indispensable contributions of both components to the effectiveness of our method.
Furthermore, substituting SGA3D with DFA3D or DFA2D demonstrates the superior performance of SGA3D, which achieves mIoU gains of 1.00 and 0.42 over DFA2D and DFA3D, respectively, while maintaining a lower parameter count and computational overhead.

\noindent\textbf{Global Similarity-Guided Attention.}
We conducted an ablation study on the GSGA module in Table~\ref{tab:SGDFA}.
Removing GSGA leads to a decline in both IoU and mIoU metrics, indicating that it effectively mitigates the information loss caused by the foreground dependency in SGA3D and alleviates the spatial constraints imposed by the prior mask.
In a separate experiment, removing similarity-guided global feature processing and replacing it with a naive global operation introduces ambiguity into both geometry and semantic features causing a degradation in overall performance.  
These results collectively validate the effectiveness of our design.

\noindent\textbf{Dynamic Instance Decoder.}
In Table~\ref{tab:instance_fusion}, we compare different fusion strategies in the Dynamic Instance Decoder. Compared to directly summing, the weighted approaches including sigmoid and softmax show better performance, verifying that different instances have different effects in fusion. Finally, by dynamically predicting the decision of each instance, our approach achieves the best performance.

\section{Conclusion}
We propose Ocean, an object-centric framework for vision-based 3D Semantic Scene Completion. Unlike ego-centric approaches, Ocean models scenes from an instance-level perspective to enhance semantic and geometric understanding. It leverages instance priors to guide object-centric feature learning through attention-based aggregation and spatial diffusion.
Ocean achieves state-of-the-art performance on the SemanticKITTI and SSCBench-KITTI-360 datasets, showing the effectiveness of the object-centric paradigm.

\section{Acknowledgments}
This work was supported in part by the National Natural Science Foundation of China under Grants 62073066 and in part by 111 Project under Grant B16009.


\bibliography{aaai2026}

\clearpage
\title{\vspace{-1cm}Supplementary Material for Ocean\vspace{0.2 cm}}

\section{A\hspace{1em}More Implementation Details}
\subsection{Dataset}
SemanticKITTI consists of $22$ driving sequences, divided into $10/1/11$ sequences for training/validation/testing. 
The dataset focuses on a 3D space of $51.2$ meters in front of the vehicle, $25.6$ meters to each side, and $6.4$ meters in height, with a voxel resolution of $0.2m$. This setup produces a prediction volume of $256 \times 256 \times 32$ voxels, with each voxel labeled into one of $20$ classes.
SSCBench-KITTI360 primarily consists of suburban scenes and includes $7$ sequences for training, $1$ for validation, and $1$ for testing, covering a total of $19$ semantic classes. The other settings are the same as SemanticKITTI.
\subsection{Metrics}
We report the intersection over union (IoU) and mean IoU (mIoU) metrics in the evaluation, which measure the geometric prediction of each voxel and semantic prediction of occupied voxels, respectively.
IoU emphasizes spatial geometry, identifying whether each voxel in a 3D space is occupied or empty. In contrast, mIoU combines both semantic and geometric information and classifies the contents of each occupied voxel. 

\subsection{Training Loss}
Following previous work, We employ $\mathcal{L}_{ce}$, $\mathcal{L}^{sem}_{scal}$, and $\mathcal{L}^{geo}_{scal}$ for semantic and geometric occupancy prediction, while $\mathcal{L}_d$ supervises the intermediate depth distribution for view transformation. Moreover, To facilitate faster convergence and enhance the quality of the BEV representation generated by slots, we incorporate the reconstruction loss  $\mathcal{L}_{recon}$. The overall loss function is formulated as follows:
\begin{align}
    \mathcal{L} = \lambda_d \mathcal{L}_d + \lambda_r\mathcal{L}_{recon}+ \mathcal{L}_{ce} + \mathcal{L}^{geo}_{scal} + \mathcal{L}^{sem}_{scal}
\end{align}
where the loss weights  $\lambda_d$ and $\lambda_r$ are set to 0.001 and 0.1, respectively.

\begin{figure*}[t]
    \centering
    \includegraphics[width= 0.8 \linewidth]{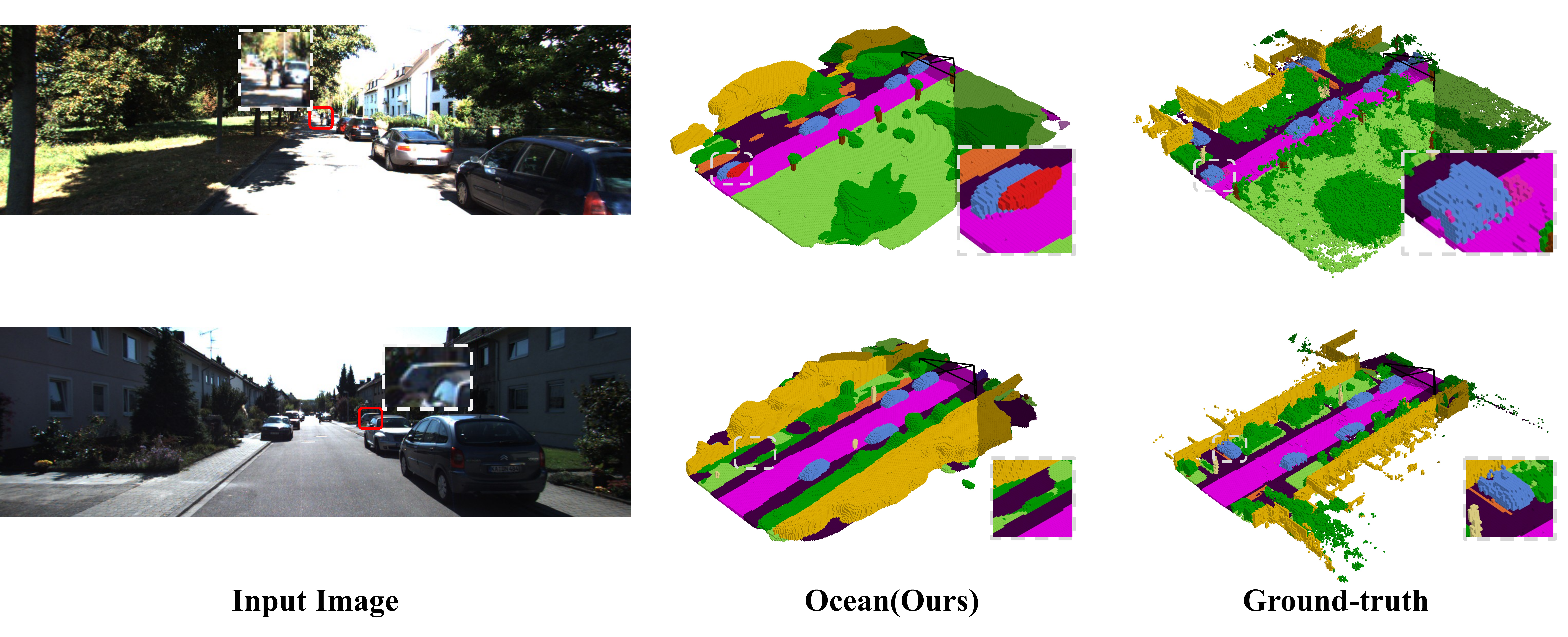}
    \caption{Failure case visualization. Insufficient visual information caused by long-range distance or severe occlusion poses a major challenge for object recognition. Under such conditions, the model often struggles to extract discriminative features, which leads to reduced recognition accuracy and increased ambiguity in classification.
    }
    \label{fig:fail}
\end{figure*}

\section{B\hspace{1em}Additional Experimental Results}
\subsection{More Exploration of Integrating MobileSAM}
In addition to SGA3D, we investigate an alternative strategy for integrating MobileSAM into the SSC task. Specifically, after the semantic grouping process, we pool the pixel features within each cluster to generate an instance feature $\mathcal{U} \in \mathbb{R}^{1\times C_2}$. This instance feature is then used as the key $K$ and value $V$ in a cross-attention mechanism alongside the proposal features $\mathcal{Q}$. We refer to this method as Semantic Instance Attention (SIA). 
Furthermore, based on the clustered features, we perform a simple feature fusion between the proposal features and the corresponding image features of the same instance, a process we refer to as Semantic Group Fusion (SGF). It is important to note that we focus solely on SGA3D in this comparison, while the remaining modules are kept consistent with our proposed method, Ocean.
We then compare these two variants with our full method and an ablated version without the SGA3D module to evaluate their relative performance.

As shown in Table~\ref{tab:sia}, SIA experiences a performance drop of 0.62/0.54 in IoU/mIoU due to the direct pooling of instance features, which results in substantial information loss. In comparison, SGF leverages all instance features and facilitates intra-instance feature interaction, thereby achieving relatively better performance. However, this improvement comes at the expense of increased parameter count and computational complexity. In contrast, our method, without relying on depth information, achieves improvements of 0.08/0.10 in IoU/mIoU while maintaining the lowest number of parameters and computational cost. 
Furthermore, with the integration of the 3D Extension, our model reaches 46.40 IoU and 17.39 mIoU. 
These results demonstrate the efficiency of our SGA3D module, which establishes a strong foundation for the object-centric paradigm.

\begin{table}[h]
\centering
\setlength{\tabcolsep}{3 pt}
{
\begin{tabular}{cc|cccc}
\toprule[.05cm]
\multicolumn{2}{c|}{\textbf{Method}} & \textbf{IoU ↑} & \textbf{mIoU ↑} & \textbf{Param} & \textbf{FLOPs}  \\
\midrule
S1       & SIA     & 45.78 & 16.85 &  0.280M  & 1.333G \\
S2       & SGF  & 45.76 & 17.20   &1.516M & 7.197G   \\
S3       & Ours(w/o Ext.) & 45.84 & 17.30  & 0.280M  & 1.333G \\
S4       & Ours   &\textbf{46.40}     & \textbf{17.39} &  0.280M  & 1.333G   \\ 
\bottomrule[.05cm]
\end{tabular}
}
\caption{Comparison of Intergating MobileSAM.}
\label{tab:sia}
\end{table}

\subsection{SemanticKITTI Validation Performance}
We show the comparison between our method and the previous method in the SemanticKITTI validation set in Table~\ref{tab:sem_kitti_val}.
Compared to other methods, our approach still achieves the best results in terms of IoU and mIoU.

%

\section{C\hspace{1em}Failure Case Analysis}
We present some representative failure cases of Ocean in Figure~\ref{fig:fail}. 
We observe that Ocean may struggle to recognize certain objects under extremely limited visual conditions, such as long distances or heavy occlusions.
In the first row of Figure~\ref{fig:fail}, although the model successfully detects the presence of an object, it fails to distinguish between a person and a bicyclist due to the large distance. Similarly, in the second row, the vehicle is heavily occluded and located far from the sensor, leading to insufficient information for accurate recognition. In future work, we plan to address these limitations by incorporating multimodal and temporal cues.

\section{D\hspace{1em}Visualization}
\subsection{More visualizations of Ocean}
\begin{figure*}[t]
    \centering
    \includegraphics[width=\linewidth]{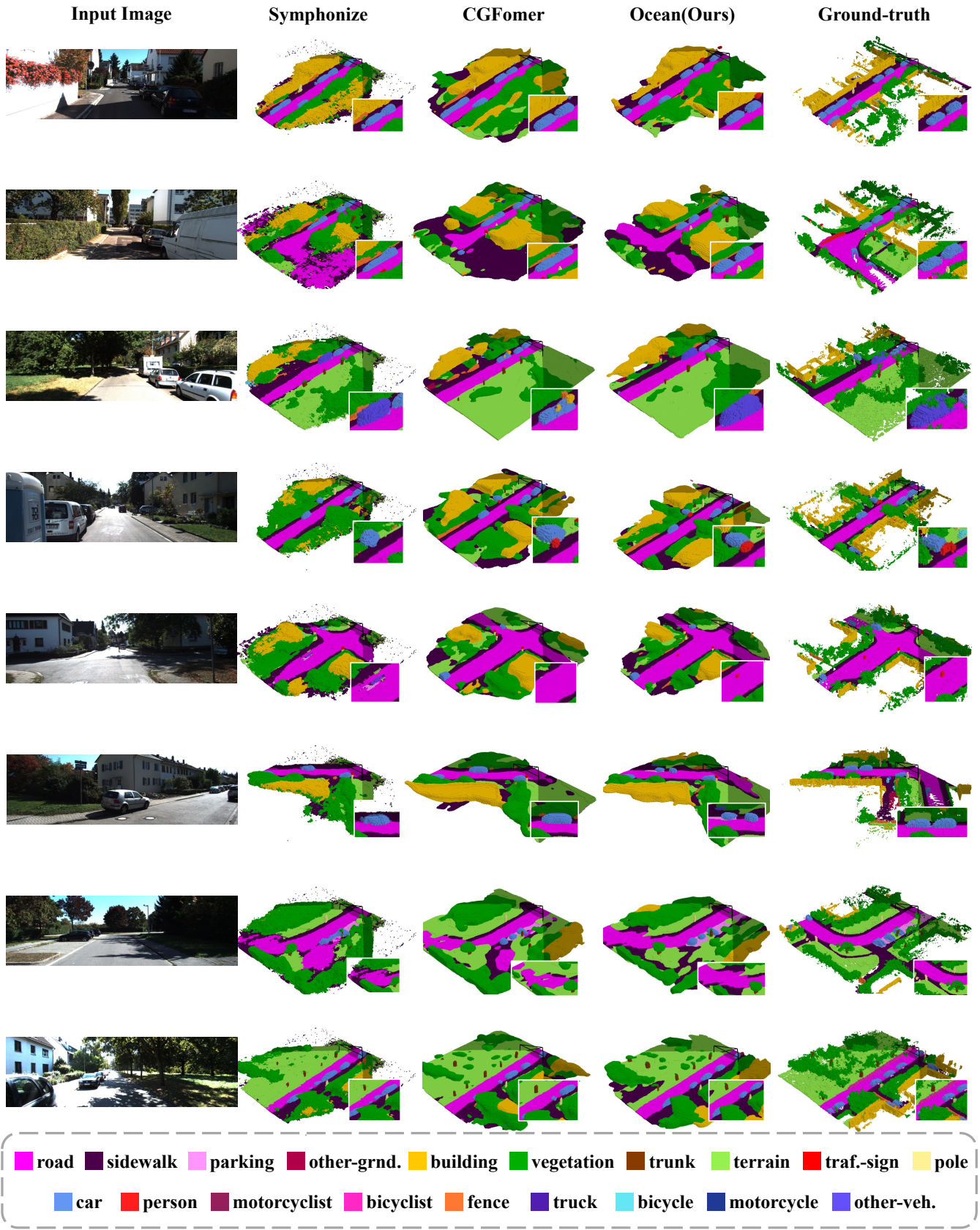}
    \caption{Qualitative visualizations on SemanticKITTI validation set. In comparison to previous methods, our method exhibits notably superior and more distinct prediction outcomes, particularly in its ability to distinguish individual instances in complex scenes.
    }
    \label{fig:supp_viz1}
\end{figure*}

We present additional qualitative results in Figure~\ref{fig:supp_viz1}, showcasing comparisons of our method with existing approaches on the SemanticKITTI validation set. The visualizations clearly demonstrate that our method excels at capturing fine-grained scene structures and delivers superior, more distinct predictions.

\begin{table*}[ht]
\centering
\newcommand{\clsname}[2]{
    \rotatebox{90}{
        \hspace{-5pt}
        \textcolor{#2}{$\blacksquare$}
        \hspace{-5pt}
        \begin{tabular}{l}
            #1                                      \\
        \end{tabular}
    }}
\setlength{\tabcolsep}{2.1pt}
{
\begin{tabular}{l|r r|rrrrrrrrrrrrrrrrrrrrr}
    \toprule[.05cm]
    Method                               &
    \multicolumn{1}{c}{IoU}              &
    mIoU                                 &
    \clsname{road}{road}                 &
    \clsname{sidewalk}{sidewalk}         &
    \clsname{parking}{parking}           &
    \clsname{other-grnd.}{otherground}   &
    \clsname{building}{building}         &
    \clsname{car}{car}                   &
    \clsname{truck}{truck}               &
    \clsname{bicycle}{bicycle}           &
    \clsname{motorcycle}{motorcycle}     &
    \clsname{other-veh.}{othervehicle}   &
    \clsname{vegetation}{vegetation}     &
    \clsname{trunk}{trunk}               &
    \clsname{terrain}{terrain}           &
    \clsname{person}{person}             &
    \clsname{bicyclist}{bicyclist}       &
    \clsname{motorcyclist}{motorcyclist} &
    \clsname{fence}{fence}               &
    \clsname{pole}{pole}                 &
    \clsname{traf.-sign}{trafficsign}
    \\
    \midrule
    MonoScene   & 36.86          & 11.08          & 56.5          & 26.7          & 14.3          & 0.5           & 14.1          & 23.3          & 7.0           & 0.6          & 0.5          & 1.5           & 17.9          & 2.81           & 29.6          & 1.9          & 1.2          & 0.0          & 5.8          & 4.1          & 2.3          \\
    TPVFormer     & 35.61          & 11.36          & 56.5          & 25.9          & 20.6          & \underline{0.9}           & 13.9          & 23.8          & 8.1           & 0.4         & 0.1          & 4.4          & 16.9          & 2.3           & 30.4          & 0.5          & 0.9          & 0.0          & 5.9          & 3.1          & 1.5          \\
    VoxFormer    & 44.02 & 12.35          & 54.8          & 26.4          & 15.5          & 0.7           & 17.7        & 25.8          & 5.6           & 0.6          & 0.5          & 3.8           & 24.4          & 5.1           & 30.0          & 1.8          & \underline{3.3} & 0.0          & 7.6          & 7.1          & 4.2          \\
    OccFormer  & 36.50          & 13.46          & 58.9          & 26.9          & 19.6          & 0.3           & 14.4          & 25.1          & \textbf{25.5} & 0.8          & 1.2          & 8.5           & 19.6          & 3.9           & 32.6 & 2.8          & 2.8          & 0.0          & 5.6          & 4.3          & 2.9          \\
    Symphonies    & 41.92          & 14.89 & 56.4          & 27.6          & 15.3          & \textbf{1.0}           & 21.6 & 28.7 & 20.4          & 2.5 & \textbf{2.8} & \underline{13.9} & 25.7 & 6.6 & 30.9           &\underline{3.5} & 2.2          & 0.0          & 8.4 & 9.6 & 5.8 \\
    CGFormer  & \underline{45.99}          & 16.87         & \underline{65.5}         & 32.3 & 20.8 & 0.2   & \underline{23.5}       & \textbf{34.3}     & 19.4   & \textbf{4.6}  & \underline{2.7}  & 7.7         & 26.9    & \underline{8.8}    & \underline{39.5} & 2.4   & \textbf{4.1}   & 0.0        & 9.2   & 10.7   & \textbf{7.8}  \\
    HTCL & 45.51         & \underline{17.13}        & 63.7          & \underline{32.5}  & \textbf{23.3} & 0.1  & \textbf{24.1}         & \textbf{34.3}    & \underline{20.7}   & \underline{4.0}  & \textbf{2.8}  & 12.0         & \underline{27.0}    & \underline{8.8}    & 37.7  & 2.6  & 2.3 & 0.0         & \textbf{11.2}  & \textbf{11.5}   &  \underline{7.0}  \\
    Ocean (Ours)  & \textbf{46.40}  & \textbf{17.39} & \textbf{66.1} & \textbf{34.3}   & \underline{21.9}   & 0.1 & 23.4 &\underline{34.0} & 19.3  & 3.2& 2.0 & \textbf{15.3}  & \textbf{28.1} &\textbf{8.9} & \textbf{40.0} & \textbf{3.7} & 1.2 & \underline{0.0} &\underline{10.3} &\underline{10.8} &\textbf{7.8} \\
    \bottomrule[.05cm]
\end{tabular}
}
\caption{Quantitative results on SemanticKITTI validation set. The best and second results are in \textbf{bold} and \underline{underlined}.}
\label{tab:sem_kitti_val}
\end{table*}

\end{document}